\documentstyle[twocolumn,psfig]{article}

\def\citi{$^\dagger$}
\def\crim{$^\ddagger$}
\def\transtalk{Trans\kern-.15em{}Talk}
\def\TT{\transtalk}
\def\argmax{\mathop{\rm argmax}}

\newcommand{\con}{\mbox{$\,|\,$}}
\newcommand{\ve}{\mbox{$\bf e$}}
\newcommand{\vf}{\mbox{$\bf f$}}
\newcommand{\vs}{\mbox{$\bf s$}}
\newcommand{\vc}{\mbox{$\bf c$}}
\newcommand{\va}{\mbox{$\bf a$}}
\newcommand{\lle}{\mbox{$|\bf e|$}}
\newcommand{\llf}{\mbox{$|\bf f|$}}

\input{psfig.tex}

\title{\large Towards an Automatic Dictation System for Translators :\\ the \TT\ Project}  
\author{\normalsize Marc Dymetman\citi\thanks{Present address: Rank Xerox Research
Centre, 6 chemin de Maupertuis, 38240 Meylan, France.}\ , Julie Brousseau\crim, George Foster\citi, \\
\normalsize Pierre Isabelle\citi, Yves Normandin\crim, Pierre Plamondon\citi \\[2mm]
\normalsize \citi\ Centre d'Innovation en Technologies de l'Information (CITI) \\
\normalsize 1575, Boul. Chomedey, Laval H7V 2X2, Quebec, Canada \\[2mm]
\normalsize \crim\ Centre de Recherche Informatique de Montreal (CRIM)\\
\normalsize 1801, McGill College, Montreal H3A 2NA, Quebec, Canada}
\date{\normalsize \today}

\begin{document}           
\begin{small}
\bibliographystyle{plain}

\nocite{%
%Church+91f,%
%foster:91,%
DymFosIsa92,%
Simard+92,%
Brown-MathsOfMT:93,%
IsabelleTMI93,%
Brown-StatisticalMT:90%
}

\maketitle              

\begin{abstract}
Professional translators often dictate their translations orally and
have them typed afterwards. The \TT\ project aims at automating the
second part of this process. Its originality as a dictation system
lies in the fact that both the acoustic signal produced by the
translator and the source text under translation are made available to
the system. Probable translations of the source text can be predicted
and these predictions used to help the speech recognition system in
its lexical choices. We present the results of the first prototype,
which show a marked improvement in the performance of the speech
recognition task when translation predictions are taken into account.
\end{abstract}

\section{Introduction}

The integration of machine translation and speech technology is
currently the focus of major projects in several countries
\cite{Kay+94,Kurematsu:87,Stentiford:90}. Usually, the aim of these
efforts is some type of speech-to-speech translation, where speech
recognition, machine translation and speech synthesis are performed
sequentially. However, both speech recognition and machine translation
are tasks that can at present be reliably accomplished only under
stringent lexical, syntactic and semantic restrictions, and
consequently developers of speech-to-speech translation systems need
to find application domains for which narrow sub-languages can be
naturally defined.

In the \TT\ project, we attempt to integrate speech recognition and
machine translation in a way which, instead of compounding the
weaknesses of both technologies, makes maximal use of their
complementary strengths.  We do not try to replace the human
translator by a machine (a hopeless endeavor, in general), but
undertake instead the more realistic task of providing a {\em
dictation tool} to the translator. Our aim is to use machine
translation to make probabilistic predictions of the possible target
language verbalizations freely produced by the translator, and to use
these predictions to reduce the difficulty of the speech recognition
task to such an extent that complete recognition of the translator's
utterances can be achieved.\footnote{The idea was independently
advanced by us \cite{DymFosIsa92} and by researchers at the IBM Thomas
J. Watson Research Center \cite{Brown-AutSpeechRecInMT:92u}.}

\begin{figure}
\centerline{\psfig{figure=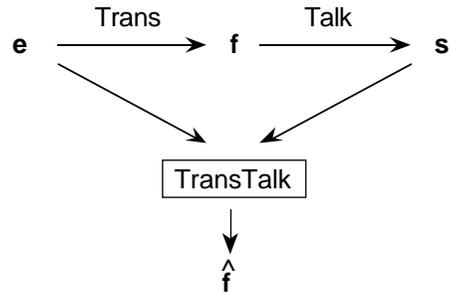}}
\caption{\protect\small \TT's underlying model. Starting from an
English sentence $\ve$, the translator mentally formulates its French
translation $\vf$, then produces its acoustic rendering $\vs$. The
system's aim is to find $\hat{\vf} = \argmax_f p(\vf \con \ve, \vs)$,
or equivalently, from Bayes's formula, $\hat{\vf} = \argmax_f p(\vs
\con \ve, \vf) \cdot p(\vf \con \ve)$.  By neglecting the influence of
$\ve$ on $\vs$ once $\vf$ is known, we can take $\hat{\vf} = \argmax_f
p(\vs \con \vf)\cdot p(\vf \con \ve)$.}
\label{triangle}
\end{figure}

For example, suppose that, in the case of English-to-French
translation, the translator decides to render the sentence ``what
splendid horses you have'' as ``tes chevaux sont vraiment
magnifiques''. A speech recognition system without access to the
source text might have difficulty distinguishing {\em chevaux}
(horses) from the acoustically close, and contextually more likely,
{\em cheveux} (hair). On the other hand, the presence in the English
source of the word {\em horses} serves as a strong indicator that the
correct choice should be {\em chevaux}, and it is on such knowledge of
probable translations that \TT\ attempts to capitalize.

Conceptually, the main difference between a conventional ``noisy
channel'' speech recognition system for French and \TT\ is that,
instead of maximizing in $\vf$ the product $p(\vs\con{}\vf) \cdot
p(\vf)$ of an ``acoustic model'' and a ``language model'' for French
(where $\vs$ stands for the acoustic signal and $\vf$ for the French
sentence), we maximize the product $p(\vs\con{}\vf) \cdot
p(\vf\con{}\ve)$ of an acoustic model and a ``translation model'' from
English to French (where $\ve$ stands for the English sentence under
translation). See figure \ref{triangle}.

We have implemented a prototype version of \TT\ that operates in an
isolated-word dictation mode over a vocabulary of 20,000 French word
forms. It is specialized for the domain of Canadian Parliamentary
debates, which are transcribed in bilingual form in the Canadian
Hansard corpus. Two years of Hansard transcripts (approximately 10M
French words and 10M English words) were used as training data for the
translation model.

\section{Acoustic model}

We use an HMM based on context-independent phone models to describe
$p(\vs \con \vf)$.
The
\TT\ vocabulary is represented with a set of 47 phonemes including 20
vowels and 27 consonnants. The base pronunciations were obtained using
a set of grapheme-to-phoneme rules which take into account phonetic
particularities found in the French spoken in Quebec such as
assibilation and vowel laxing. 

Recognition is performed with an $n$-best search of a
compressed phonetic graph representing the entire 20,000 word
vocabulary \cite{Lacouture+91}. This graph is such that no two paths
produce the same phone sequence and every path corresponds to a valid
phonetic representation in the dictionary. A given path will therefore
correspond to all lexicon entries sharing the same phonetics.  
The search yields a list containing the 20 most acoustically probable
words for each (isolated) acoustic token.

\section{Translation model}

The aim of the translation model is to describe $p(\vf \con \ve)$, the
probability that a translator will produce a French translation \vf\
for an English sentence \ve. 

\subsection{Modelling Approaches}

There are at least two distinct approaches to modelling this
distribution. In \cite{Brown-MathsOfMT:93}, Brown et al.\ expand it as
the product $p(\vf) \cdot p(\ve \con \vf)$, to which it is
proportional under maximization over \vf. The main advantage of this
arrangement is that it provides for a division of labour in which
$p(\vf)$ is responsible for the well-formedness of \vf, and $p(\ve
\con \vf)$ for ensuring that \ve\ and \vf\ are acceptable translations
without having to be unduly preoccupied with the internal structure of
either.  Although this is a powerful technique, it has one drawback
that makes it unsuitable for our purposes: it does not easily lend
itself to efficient searches over large sets of French sentence
candidates.

Because of this, we have chosen to model $p(\vf \con \ve)$ more
directly as a family of parameterized Markov language models
$p_{\lambda(e)}(\vf)$, where each \ve\ specifies a parameter vector
$\lambda$, not necessarily uniquely.  This approach presents the
challenge of incorporating information from \ve\ in a way that does
not interfere with the language model's knowledge of the structure of
French---particularly for language models that are accurate to begin
with.  In the work reported here we have largely avoided this
difficulty by using a fairly weak language model; our aim is mainly to
investigate to what exent the performance of such a model can be
improved without substantially increasing its low run-time cost.

\subsection{Derivation}

The translation model is based on a standard tri-class language model
conditioned on \ve. The first key assumption we make is that the
sequence \vc\ of word classes for \vf\ is independent of \ve, which
allows us to write:
\begin{equation}
   p(\vf \con \ve) = \sum_{\vc} p(\vc) \cdot p(\vf \con \vc, \ve) 
\end{equation}
This approximation is motivated by the intuition that \ve\ will be
most informative about the actual words in \vf, and only weakly
informative about gross syntactic structure of the sort that \vc\
captures. Because it is most valid when \vc\ consists of broad
classifications\footnote{This assumption becomes increasingly
untenable for finer classification schemes; in the limit when classes
are identical to words, the model collapses into a pure tri-gram with
no translation component whatsoever.}  we use a minimal set of 15
classes which correspond to the major grammatical categories (noun,
verb, etc).

To incorporate translation information, we suppose, following Brown et
al.\ \cite{Brown-MathsOfMT:93}, that \vf\ and \ve\ are related via an
{\em alignment} (see figure~\ref{align}) in which each French word is
connected to either a single English word in \ve\ or none at all. An
alignment can be represented as a vector \va\ of length \llf\ which
contains, for each French word, the position in \ve\ of the English
word to which it connects, or zero if it is not connected. 
We assume
that all $A_{\vf,\ve}$ possible alignments are equally likely,
with probability $p(\va \con \vc, \ve) = 1/A_{\vf,\ve}$, so we
have:\footnote{Where $A_{\vf,\ve} = (\lle + 1)^{\llf}$}.
\begin{equation} \label{sumeq}
   p(\vf \con \vc, \ve) = \sum_{\va} \frac{1}{A_{\vf,\ve}} \cdot p(\vf \con \va, \vc, \ve) 
\end{equation}
This is a rough approximation which runs contrary to our knowledge
that some alignments---such as those in which all French words connect
to a single English word, or those in which French verbs connect only
to English prepositions---will be much less likely than others. Its
purpose is simplification, and we justify it on the grounds that a
reasonable model for $p(\vf \con \va, \vc, \ve)$ will minimize the
contribution from (most) poor alignments in any case.

\begin{figure}
\centerline{\psfig{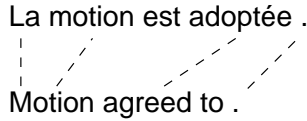}}
\caption{\protect\small An example of an alignment, one of
$5^5$ which are possible for this sentence pair.}
\label{align}
\end{figure}

\begin{figure}
\centerline{\psfig{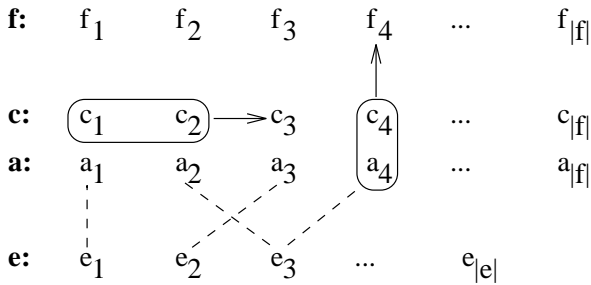}}
\caption{\protect\small The structure of the Markov source underlying
the translation model.  First, \vc\ is established by choosing each
class based on the previous two with probability given by the
appropriate contextual parameter. Next \va\ is established by picking
a position in \ve\ at random for each position in \vc. Finally, \vf\
is generated by choosing each word based on its class and its English
partner, with probability given by the appropriate bi-lexical
parameter.}
\label{paralleles}
\end{figure}

The final step is to assume that the words in \vf\ are conditionally 
independent
given \va, \vc, and \ve, and furthermore that each word depends only on its
class and the English word to which it connects in the alignment:
\begin{equation}
   p(\vf \con \va, \vc, \ve) = \prod_{i=1}^{\llf} p(f_i \con c_i, e_{a_i})
\end{equation}

Our complete model is a Markov source (see figure~\ref{paralleles}) which
depends on two sets of parameters: {\em contextual} parameters of the
form $p(c_i \con c_{i-2}, c_{i-1})$, which predict a class from its
two predecessors; and {\em bi-lexical} parameters of the form $p(f
\con c, e)$, which predict a French word from its class and its
English partner.

It is possible to rearrange the straightforward combination of
equations 1, 2, and 3 in a way which permits more efficient
calculations. The key observation is that the sum over all alignments
can be reorganized into a product of sums over English words. The
result is the equation
\begin{equation} \label{main}
   p(\vf \con \ve) = \sum_{\vc}
                     \prod_{i=1}^{\llf} p(c_i \con c_{i-2}, c_{i-1})
                     p(f_i \con c_i, \ve)
\end{equation}
where $p(f_i \con c_i, \ve) = \sum_{j=0}^{\lle} p(f_i \con c_i, e_j) /
(\lle + 1)$. From this it should be obvious that our translation model
is nothing more that a standard tri-class model in which the lexical
parameters $p(f \con c)$ have been replaced by $p(f \con c, \ve)$.

\subsection{Parameter Estimation} \label{estm}

The two families of parameters in the translation model were estimated
separately. Contextual parameters were estimated as part of a pure
tri-class language model for French, which was trained on the French
half of our bilingual corpus via the EM algorithm, using a
dictionary to identify valid classes for each word.

Bi-lexical parameters were estimated as part of a simplified
translation model in which contextual information was assumed to
be explicit:
\begin{equation} \label{simp_trans}
   p(\vf, \vc \con \ve) = \frac{1}{A_{\vf,\ve}} \sum_{\va} \prod_{i=1}^{\llf} p(f_i, c_i|e_{a_i})
\end{equation}
To train this model, we first aligned the training corpus to the
sentence level using the method described in \cite{Simard+92}. To
improve the quality of our training data, we filtered out alignments
which involved more than one sentence in either language as well as
those which contained more than 40 tokens in either language---this
reduced the size of the training set by approximately 20\%, to about
8M tokens in each language. Next, we used the pure language model to
tag each word in the French part of the reduced corpus with its most
likely class. Finally, we used the EM algorithm to estimate parameters
$p(f, c \con e)$ from the aligned, tagged corpus. These were transformed
into bi-lexical parameters as follows:
\begin{equation}
   p(f \con c, e) = \frac{p(f, c \con e)}{\sum_f p(f, c \con e)}
\end{equation}
Figure~\ref{trans-sample} shows a sample of the results.

\begin{figure} \centering
\begin{tabular}{|l|l|} 
\hline
$f$		& $p(f \con c, e)$ \\
\hline
gouvernement	& 0.7363 \\
m.		& 0.0227 \\
monsieur	& 0.0134 \\
président	& 0.0109 \\
canada		& 0.0081 \\
façon		& 0.0033 \\
mesure		& 0.0024 \\
part		& 0.0023 \\
ministre	& 0.0023 \\
décision	& 0.0022 \\
\hline
\end{tabular}
\caption{\protect\small A sample of \TT's bi-lexical parameters. These
are the ten most probable French words, given the class NOUN and
the English word {\em government}.}
\label{trans-sample}
\end{figure}

Because many valid bi-lexical combinations do not occur in our
training corpus, it was necessary to smooth the bi-lexical
parameters. Rather than modifying the empirical distribution $p(f \con
c, e)$ directly, we chose to dynamically smooth the more robust
quantity $p(f \con c, \ve)$ involved in calculations based on
equation~\ref{main}.  We experimented with three simple methods of
combining this with the less precise but more reliable lexical
parameters $p(f \con c)$ from the pure language model: linear
interpolation; using the maximum of $p(f \con c, \ve)$ and $p(f \con
c)$; and using $p(f \con c)$ iff $\max_e p(f \con c, e) / p(f \con c,
\ve)$ did not exceed some threshold. The rationale for the second
method is that we expect higher probabilities to be more
reliably estimated on average than lower ones. The third method 
is intended to
reject translation information when there is no English word that is
strongly associated with the current French word. Because the last two
methods result in unnormalized distribitions, they can be compared
only in terms of recognition performance and not by 
means of the perplexity
measure (see section~\ref{results}).

\section{Search}

The aim of the search component is to find 
an approximation to
the sentence $\hat{\vf}$ that maximizes
the product of acoustic and translation scores $p(\vs \con
\vf) \cdot p(\vf \con \ve)$. Our search algorithm is divided 
into two stages, both of
which are suboptimal.

The first stage involves using the acoustic model to prune the list of
word hypotheses for each acoustic token from 20,000 to some number $n$
(currently 20). Since this pruning is performed without reference to
the translation model, there is no guarantee that $\hat{\vf}$ is among
the $n^{\llf}$ sentence candidates retained.

The second stage is a Viterbi search through the remaining sentence
candidates using the translation model.  This permits us to find the
pair $(\tilde{\vf}, \tilde{\vc})$ that maximizes the product $p(\vs
\con \vf) \cdot p(\vf, \vc \con \ve)$ in time which is proportional to
$n C^3 \llf \lle$, where $C$ is the number of word classes in the
translation model (currently 15). Given the coarse nature of our word
classes, we feel that $\tilde{\vf}$ is a reasonable approximation to
$\hat{\vf}$.

\section{Results} \label{results}

We tested \TT\ on a small corpus of 50 French/English sentence pairs
from the Hansard corpus which were not used as training data. The
French sentences were all between 15 and 20 tokens in length (counting
punctuation) and were selected so as not to contain words outside our
20,000 word vocabulary. They were dictated in isolated-word mode by
two different speakers.

Figure~\ref{exa} illustrates the results for a single sentence pair.
Overall statistics are given in figure~\ref{stats}.  The translation
model yielded an average error-rate decrease of 24\% over the pure
language model. For errors which involved ``content'' words (eg, {\em
action} for {\em section}) the decrease was 42\%. The perplexity of
the test corpus was reduced by more than half by the use of the
translation model.

\begin{figure}
\centerline{\psfig{figure=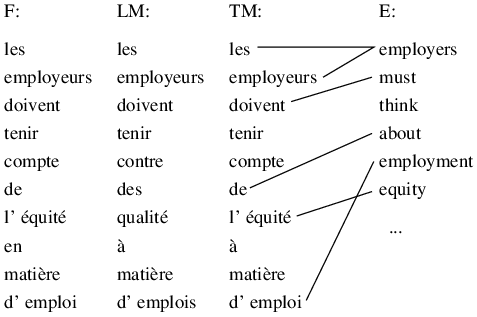}}
\caption{\protect\small Comparison of language model (LM) and
translation model (TM) results for a sentence pair (F,E) from the test
corpus. (This pair has been truncated for space reasons.) Lines
indicate salient parts of the most probable alignment between the
output sentence and E. The presence of {\em equity} in the English
source allowed the translation model to correctly choose {\em équité}
instead of {\em qualité}. }
\label{exa}
\end{figure}

\begin{figure} 
\centering
\begin{tabular}{|l|l|l|l|}
\hline
Model	& \multicolumn{2}{c|}{Words Correct (/918)} & Perplexity \\
\cline{2-3}
	& Speaker 1 & Speaker 2 & \\
\hline
pure language		& 686 (74.7\%) & 677 (73.8\%) & 385 \\
interpolated (.85)	& 735 (80.1\%) & 734 (80.0\%) & 180 \\
maximum			& 735 (80.1\%) & 732 (79.7\%) & -- \\
e-testing (.30)		& 742 (80.8\%) & 734 (80.0\%) & -- \\
\hline
\end{tabular}
\caption{\protect\small Summary of \TT\ results. The first line
contains statistics for the pure language model; the remaining lines
contain statistics for the translation model with each of the three
different smoothing methods described in section \protect\ref{estm}
(where $.85$ was the optimum the weighting factor for bi-lexical
parameters, and $.30$ was the optimum confidence threshold).}
\label{stats}
\end{figure}

\section{Conclusions}

Our initial results demonstrate that it is possible to cheaply and
effectively make use of translation information for speech
recognition. We feel that the simple approach described in this paper
barely begins to tap the potential of the \TT\ idea, and we are
currently investigating a number of ways of improving on it.

\end{small}
\end{document}